\documentclass[10pt,journal,compsoc]{IEEEtran}\usepackage[]{graphicx}\usepackage[]{color}
\makeatletter
\def\maxwidth{ %
  \ifdim\Gin@nat@width>\linewidth
    \linewidth
  \else
    \Gin@nat@width
  \fi
}
\makeatother

\definecolor{fgcolor}{rgb}{0.345, 0.345, 0.345}

\usepackage{framed}
\makeatletter
 {\par\unskip\endMakeFramed%
 \at@end@of@kframe}
\makeatother

\definecolor{shadecolor}{rgb}{.97, .97, .97}
\definecolor{messagecolor}{rgb}{0, 0, 0}
\definecolor{warningcolor}{rgb}{1, 0, 1}
\definecolor{errorcolor}{rgb}{1, 0, 0}

\usepackage{alltt}

\usepackage{censor}
\usepackage{booktabs}

\ifCLASSOPTIONcompsoc
  \usepackage[nocompress]{cite}
\else
  \usepackage{cite}
\fi

\ifCLASSINFOpdf
\else
\fi

\usepackage{amsmath}

\usepackage{url}

\IfFileExists{upquote.sty}{\usepackage{upquote}}{}
\begin{document}
%

\title{Reliability and Validity of Image-Based and Self-Reported Skin Phenotype Metrics}

%
%
%

\author{John~J.~Howard,
        Yevgeniy~B.~Sirotin,
        Jerry~L.~Tipton,
        and~Arun~R.~Vemury 
\IEEEcompsocitemizethanks{\IEEEcompsocthanksitem J. Howard, Y. Sirotin, and J. Tipton work at the Maryland Test Facility in Upper Malboro, Maryland.  \protect\\
\IEEEcompsocthanksitem A. Vemury works at the United States Department of Homeland Security, Science and Technology Directorate in Washington, DC.\protect\\
\IEEEcompsocthanksitem Authors listed alphabetically. E-mail correspondence should be sent to info@mdtf.org}}

%
%

\markboth{Pre-print Copy}%
{Shell \MakeLowercase{\textit{et al.}}: Bare Advanced Demo of IEEEtran.cls for IEEE Biometrics Council Journals}
%


\IEEEtitleabstractindextext{%
\begin{abstract}
With increasing adoption of face recognition systems, it is important to ensure adequate performance of these technologies across demographic groups, such as race, age, and gender.  Recently, phenotypes such as skin tone, have been proposed as superior alternatives to traditional race categories when exploring performance differentials.  However, there is little consensus regarding how to appropriately measure skin tone in evaluations of biometric performance or in AI more broadly.  Biometric researchers have estimated skin tone, most notably focusing on face area lightness measures (FALMs) using automated color analysis or Fitzpatrick Skin Types (FST). These estimates have generally been based on the same images used to assess biometric performance, which are often collected using unknown and varied devices, at unknown and varied times, and under unknown and varied environmental conditions.  In this study, we explore the relationship between FALMs estimated from images and ground-truth skin readings collected using a colormeter device specifically designed to measure human skin.  FALMs estimated from different images of the same individual varied significantly relative to ground-truth FALMs. This variation was only reduced by greater control of acquisition (camera, background, and environmental conditions). Next, we compare ground-truth FALMs to FST categories obtained using the standard, in-person, medical survey.  We found that there was relatively little change in ground-truth FALMs across different FST category and that FST correlated more with self-reported race than with ground-truth FALMs.  These findings show FST is poorly predictive of skin tone and should not be used as such in evaluations of computer vision applications.  Finally, using modeling, we show that when face recognition performance is driven by FALMs and independent of race, noisy FALM estimates can lead to erroneous selection of race as a key correlate of biometric performance. These results demonstrate that measures of skin type for biometric performance evaluations must come from objective, characterized, and controlled sources. Further, despite this being a currently practiced approach, estimating FST categories and FALMs from uncontrolled imagery does not provide an appropriate measure of skin tone.
\end{abstract}

}

\maketitle

\IEEEdisplaynontitleabstractindextext

%
\IEEEpeerreviewmaketitle

\ifCLASSOPTIONcompsoc
\IEEEraisesectionheading{\section{Introduction}\label{sec:introduction}}
\else
\section{Introduction}
\label{section:introduction}
\fi

\IEEEPARstart{B}{iometric} technologies are increasingly being adopted for use as a means of asserting identity in banking, medicine, travel, and a variety of government applications.  As reliance on biometrics increases, it is important to demonstrate that these technologies are not only accurate, but also fair, i.e. that they are consistently accurate for different groups of people.  Many factors can contribute to differences in accuracy across groups, such as algorithm architecture, training image properties, biometric properties, training set composition, test image properties, and individual behavior~\cite{suresh2019framework}.  Thus, it is important to test candidate biometric systems to quantify any differences between groups to help determine whether they are fair.  However, the process of dividing individuals into different categories or groups for evaluation can be problematic.

Face recognition is a type of biometric that can identify a human individual by using the unique physiological features of their face.  Previous studies of fairness in face recognition have divided individuals into demographic groups including gender and race~\cite{howard2019effect, vangara2019characterizing, grother2019face3}.  However, grouping individuals based on social categories has several drawbacks.  First, social categories evolve over time, causing some individuals to shift group membership and causing some groups to disappear altogether.  Second, different regions can have different social categories and social category definitions. Finally, social category definition sets are not guaranteed to reflect the range of variation of face physiological features in a population~\cite{ma2018race}.

For these reasons, recent studies have proposed relating performance to phenotypic measures as a more scientifically useful analysis of algorithm fairness~\cite{cook2019demographic, buolamwini2018gender}.  Phenotypes are observable characteristics of a person and as such may offer better explanations of any observed biometric performance variation.  However, techniques for assigning phenotypes to individuals are currently understudied.  In~\cite{buolamwini2018gender}, images were taken from government websites of three African countries and three European countries and manually assigned a numerical value inspired by the Fitzpatrick Skin Type (FST) categories.  Since then, FST has been proposed as a measurement of relevance in additional face biometric studies~\cite{muthukumar2018understanding, krishnapriya2020issues, lu2019experimental}, studies of fairness in self-driving car algorithms~\cite{wilson2019predictive}, and even proposed as a common benchmark for detailing the performance characteristics of machine learning algorithms generally~\cite{mitchell2019model}.

If FST is to become a consensus measure of relevance in machine learning more broadly and in fairness studies specifically, it's appropriate to scrutinize both its use and the way in which it has been measured in previous studies.  A good measure of skin type should be 1) consistent and 2) representative of the underlying phenotype.  Unfortunately, the degree to which skin type measured from images meets these two criteria has not been well assessed and there are reasons to question the efficacy of this practice.  For example, face skin tone in a photograph might be hard to distinguish from the amount of light illuminating the face (due to variation in face pose and ambient illumination when the photograph was taken) and from variation in camera settings (e.g. aperture, shutter speed, ISO).  Also, in regards to FST, existing behavioral literature has found humans do not always accurately determine FST~\cite{harrison1999all}~\cite{reeder2010questionnaire}~\cite{hill2002race}~\cite{krishnapriya2021analysis}.  Finally, the fundamental appropriateness of FST as a skin type metric has not been explored in the context of computer science tasks, despite documented concerns from the medical community as to the effectiveness of the FST measure~\cite{westerhof1990relation, leenutaphong1995relationship, pichon2010measuring, galindo2007sun, sommers2019fitzpatrick}.

In this study, we introduce the term Face Area Lightness Measures (FALMs) to be any technique for characterizing the intensity of light reflected by human skin in the facial region, as measured by a sensor (this has been called many things in previous studies: lighter/darker-skin~\cite{buolamwini2018gender}, tone~\cite{krishnapriya2020issues, lu2019experimental, wilson2019predictive}, reflectance~\cite{cook2019demographic}, etc.).  We assess variation in FALMs estimated from images taken in various environments, at various times, and on various devices and compare these measures to ground-truth measurements from a calibrated dermological device, designed specifically to measure skin lightness.  We then explore the suitability of FST as a proxy to FALMs by comparing ground-truth FALM readings with subjects' self-reported FST.  Finally, we perform data simulation and modelling to show that poor FALM estimations can result in the erroneous selection of categorical demographic race as the significant explanatory cause of performance variation, even when this variation is primarily driven by FALMs.


\section{Background}
\label{section:Background}

\subsection{Demographics and biometric performance}
\label{subsection:DemographicPerformance}

Prior studies have examined differences in biometric performance across demographic groups. For instance, face recognition algorithms trained on one demographic were found to perform better on that demographic~\cite{klare2012face}. Another study noted that some face datasets used for algorithm development under-represented people with darker skin and that gender classification algorithms had poorer performance for women with darker-skin~\cite{buolamwini2018gender}.  However, it is not clear whether skin tone and race can be used interchangeably and are equally related to biometric performance or indeed contribute equally to different measures of biometric performance.  For example, African-American subjects were shown to have higher false match rate (FMR) relative to Caucasians, but the difference was not due to variation in skin tone estimated from images used in matching analysis~\cite{krishnapriya2020issues}.  On the other hand, African-American subjects in another study had higher false non-match rates (FNMR) than White subjects, but this difference was better explained by poor quality acquisition of subjects with relatively lower skin reflectance, an effect not uniformly observed across cameras~\cite{cook2019demographic}.


\subsection{Optical properties of skin and skin type}
\label{subsection:SkinOptics}

Measurement of skin optical properties depends on the degree to which skin reflects, absorbs, scatters, and transmits incident light~\cite{diffey1980ultraviolet}.  Skin is a heterogeneous surface and is affected by variable amount of blood irrigation and pigmentation. Three layers of skin are visible from the surface: epidermis, dermis, and variable amounts of subcutaneous adipose tissue. The living part of the epidermis is the location of most skin pigmentation, which is caused by variable numbers of red/yellow phaeomelanin and brown/black eumelanin. In the dermis, blood, hemoglobin, beta carotene and bilirubin can absorb light, while the fibrous structure of the dermis produces scattering. Skin erythema contributes to skin redness and is related to the dilation of blood vessels nearest to the surface of the skin~\cite{diffey1980ultraviolet,van2013objective}.

Spectroscopic analysis of skin under controlled conditions allows accurate determination of the constituent chromophores.  In clinical practice, spectroscopy is routinely used to estimate melanin and erythema content using calibrated colormeter devices created for this purpose~\cite{van2013objective, clarys2000skin}.  In addition to readings of melanin and erythema, these devices provide $sRGB$ color that can be converted to the $L^{*}a^{*}b^{*}$ colorspace where lightness is represented by the L* parameter~\cite{weatherall1992skin}.  Under well controlled laboratory conditions, such readings have also been demonstrated as possible using $RGB$ cameras~\cite{nishidate2008visualizing,everett2012making}.

\subsection{Fitzpatrick skin type classification}

The Fitzpatrick Skin Type (FST) is the most used skin classification system in dermatology~\cite{ware2020racial}. The FST was originally designed to classify the UV-sensitivity of individuals with white skin for determining doses of ultraviolet A phototherapy, a treatment for dermatitis and other skin disorders.  The original FST instrument was released in 1975 and included four skin types (I-IV).  It was updated in 1988, adding two additional skin types to account for individuals with darker skin (V-VI)~\cite{fitzpatrick1988validity}.

According to medical literature, there are two ways of establishing the FST of an individual: self-report or third-party direct assessment by an expert, both are subjective and involve recording the subject's answers to questions about skin responses to sun exposure.  Fitzpatrick Skin Type was initially described as self-reported only~\cite{fitzpatrick1988validity}.  In later studies, doctors estimated FST after an in-person inspection~\cite{ware2020racial}~\cite{he2014self}.  However, even with access to the physical subject, the FST system is known to be generally unreliable estimator of skin pigmentation~\cite{westerhof1990relation} and FST types are known to be specifically less reliable for non-White individuals~\cite{leenutaphong1995relationship, pichon2010measuring, galindo2007sun, sommers2019fitzpatrick}.  Interestingly, physician-assessed FST types have been demonstrated to correlate with race, but when FSTs are self reported, the relationship between FST and race is not consistent~\cite{he2014self, chan2005assessing, ash2015evaluation}.  Indeed, recent work has pointed out that the FST in medicine, in addition to measuring skin reactivity to ultraviolet illumination, is now also also sometimes used as a proxy for race, which confounds the interpretation of the measure~\cite{ware2020racial}.  Some medical researchers have even argued against \textit{any} subjective assessments of skin type, favoring a more quantitative approach using calibrated spectrophotometers or digital cameras~\cite{everett2012making}.

In 2018, FST was utilized, for the first time, as a proxy for darker/lighter-skin in the evaluation of a computer algorithm.  That paper discussed the accuracy of gender classification algorithms across FST groups derived from face photos~\cite{buolamwini2018gender}.  This spurred numerous other computer science papers where the relationship between FST measures and performance was measured in domains such as face recognition~\cite{muthukumar2018understanding, krishnapriya2020issues, lu2019experimental, krishnapriya2021analysis} and pedestrian detection for self driving cars~\cite{wilson2019predictive}.  FST has even been proposed as a standardized method for documenting the performance of a generic machine learning algorithm~\cite{mitchell2019model}.

Crucially, the FST measures in~\cite{buolamwini2018gender, krishnapriya2020issues} and~\cite{wilson2019predictive} were determined by third-party assessment of previously acquired \emph{images} of individuals.  No direct assessment or self-report was performed as part of these studies, despite being the documented method of arriving at an FST classification, as laid out in the medical literature~\cite{fitzpatrick1988validity}~\cite{ware2020racial}~\cite{he2014self}.  Additionally, these studies did not attempt to validate that their third-party, remote assessments of FST were accurate representations of actual FST measures and did not address the well documented concerns of the medical community with the FST approach to skin color classification~\cite{westerhof1990relation, leenutaphong1995relationship, pichon2010measuring, galindo2007sun, sommers2019fitzpatrick}.  It is well established that human perception of face color is known to be affected by race~\cite{levin2006distortions} and by the color of other face features, such as lips~\cite{hayward2013other}.  Furthermore, any accurate assessment off skin type from a photograph would depend critically on the camera system and degree to which skin tone in images is represented reliably.  This representation is affected by pose, ambient lighting, choice of camera, and likely many more factors.  The concerns with third-party assessment of FST were supported by a 2021 study measured the consistency of FST ratings across different people and against automated measures, finding notable variation and inconsistencies between raters for a single image and between automated measures from different images of the same person~\cite{krishnapriya2021analysis}.

Finally, it is important to note that, when using ordinal scales, like FST, in scientific studies, altering either the survey instrument or the assigned categories alters the scale.  FST refers to a 6 point scale, arrived at by asking specific questions related to UV sensitivity~\cite{eilers2013accuracy,he2014self,ash2015evaluation,sommers2019fitzpatrick}.  Other scales, such as the IARPA IJC-B skin tone descriptions~\cite{whitelam2017iarpa}, are not described as FST, despite also being a 6 point scale.  This is because both the categories and the survey instrument are different.  Using different category labels or a different method for arriving at these labels are unlikely to produce the same ratings as the other instruments.  When new ordinal scales are introduced, care must be taken to explain both how this scale was developed and how it validated against the underlying phenomena the scale is measuring.

\section{Methods}
\label{section:Methods}

$L^{*}a^{*}b^{*}$ colorspace, particularly the L*, or lightness, component has been proposed a quantitative means for the communication of skin-color information.  It is advantageous over other colorspace representations because changes in the L* dimension relate directly to changes in human perception~\cite{weatherall1992skin}.  In this study we leverage the $L^{*}a^{*}b^{*}$ colorspace, and refer to different approaches to characterize light reflected by the skin, from the facial region and measured by a sensor, as Face Area Lightness Measures (FALMs).

We first outline the source of our subject and image data (Section~\ref{subsection:sources}) and arrange these data into various datasets that vary by level of control (Section~\ref{subsection:datasets}).  Then, to establish the consistency and appropriateness of different FALM techniques, we compare FALMs estimated from images in these datasets (Section~\ref{subsection:imagelightness}) to ground-truth FALMs collected by a calibrated instrument (Section~\ref{subsection:colormeter}).  Ground-truth FALMs are also compared to self-reported FST categories (Section~\ref{subsection:fst}).

\subsection{Sources of subject and image data}
\label{subsection:sources}

Data for this study came from two sources.  Fist, the Maryland Test Facility (MdTF) is a biometrics research lab affiliated with the U.S. Department of Homeland Security (DHS) that has been in operation since 2014.  As part of biometric technology evaluations at the MdTF, human subjects are recruited as test participants from the general population.  In particular, a test in May of 2019, acquired face photographs from 345 human subjects on nine different acquisition devices~\cite{howard2019results}.  These photographs (``Acquisition Images'') were compared to other face photographs that had been collected for each subject over a period of 1-5 years preceding the 2019 test (``Historic Images'').  Historic images were collected on a variety of different face biometric acquisition devices at the MdTF.  Acquisition Images from the nine acquisition devices were also compared to high-quality face photographs, captured by a trained operator, using a Logitech C920 webcam (``Enrollment Images'').  Enrollment Images were captured in front of a neutral grey background in accordance with ISO/IEC 19794-5. All race information from MdTF subjects was self reported by the subjects upon study enrollment.  Also, as part of~\cite{howard2019results} ground-truth measures were taken using a calibrated dermographic instrument, specifically designed to measure skin (see Section~\ref{subsection:colormeter}).

The second data source of images for this study is Special Database 32 - Multiple Encounter Dataset (``MEDS Images'') from the U.S. National Institutes of Science and Technology (NIST).  The MEDS dataset consists of mugshot photos from individuals who have had multiple encounters with law enforcement~\cite{founds2011nist}.  Race information from these subjects is included as part of the MEDS dataset but was assigned by a third party, not self reported by the subject.  Table~\ref{table:sources} summarizes the number of subjects and images for each data source used in this study.

\begin{table}
    \centering
    \begin{tabular}{l|l|r|r}
    \hline
    Source             & Race & Images & Subjects\\
    \hline
    MEDS Images        & B    & 595     & 184\\
    MEDS Images        & W    & 458     & 229\\
    \hline
    Historic Images    & B    & 1874    & 181\\
    Historic Images    & W    & 1710    & 164\\
    \hline
    Acquisition Images & B    & 1458    & 181\\
    Acquisition Images & W    & 1320    & 164\\
    \hline
    Enrollment Images  & B    & 181     & 181\\
    Enrollment Images  & W    & 164     & 164\\
    \hline
    \end{tabular}
    \caption{Images and subjects per source examined in this study.}
	\label{table:sources}
\end{table}

\begin{table*}[ht]
    \centering
    \begin{tabular}{r|r|l|l|l|l|l|l}
    \hline
    & Dataset (A) & Source                  & Image  & Environment (E) & Time (T) & Device (D) & Face Area             \\
    &			  &					        & Based  &                 &          &            & Lightness Measure ($L_f$)               \\
    \hline
    i   & MEDS        & MEDS                    & Yes & Varied          & Varied   & Varied     & $L_f$                            \\
    ii  & CE          & Historic \& Acquisition & Yes & Constant        & Varied   & Varied     & $L_f$                            \\
    iii & CET         & Acquisition             & Yes & Constant        & Constant & Varied     & $L_f$                            \\
    iv  & CED         & Historic \& Acquisition & Yes & Constant        & Varied   & Controlled & $L_{f,d}-\mu_{f,d}+\mu_f$        \\
    v   & CEDT        & Acquisition             & Yes & Constant        & Constant & Controlled & $L_{f,d}-\mu_{f,d}+\mu_f$        \\
    vi  & Corrected   & Enrollment              & Yes & Constant        & Constant & Constant   & $(L_{f,d}-L_{b,d}) + \frac{1}{2}(\mu_{f,d}-\mu_{b,d})$ \\
    vii & Ground-truth  & Colormeter              & No & Constant        & Constant & Constant   & $\frac{1}{2}(L_{rc}+L_{lc})$      \\
    \hline
    \end{tabular}
    \caption{Face area lightness datasets examined.}
	\label{table:datasets}
\end{table*}

\subsection{Face area lightness datasets}
\label{subsection:datasets}

To study the effect of various controls during the photographic acquisition process on FALMs, the data described in Section~\ref{subsection:sources} were arranged into seven distinct datasets.  Each of these datasets afforded different levels of control for environment, capture time, and device.  For each of these datasets, FALMs were calculated based on the information available.  Table~\ref{table:datasets} shows the seven face area lightnes datasets used in this study and the corresponding FALM ($L_{f}$) equations.  Sections~\ref{subsection:imagelightness} and~\ref{subsection:colormeter} outline the techniques used to calculate the FALM values for each dataset.

\subsection{FALMs from images}
\label{subsection:imagelightness}

To assess FALMs from images, pixels falling on the skin of the face were selected by face finding, circular masking, and outlier removal using methods adapted from~\cite{taylor2014adaptive} and previously used by~\cite{cook2019demographic}. The $sRGB$ values of face skin pixels were averaged and converted from $sRGB$ to the $L^{*}a^{*}b^{*}$ colorspace using the D65 illuminant.  FALMs $(L_f)$ were estimated from the resulting L* channel.

FALMs from the MEDS dataset images came from varied environment, varied devices, and varied acquisition times.  The CE and CET datasets consist of images collected in the constant environment of the MdTF, i.e. a single location with controlled/standard office lighting (600 Lux). The CET dataset consisted of images collected in a single day, i.e. constant time, ruling out any variation in subjects' actual skin pigmentation across these images.  For images in the MEDS, CE, and CET datasets there is no way to normalize $L_f$ values further and the FALM for these datasets are calculated as described in the previous paragraph.

However, the CED and CEDT datasets consist of images collected at the MdTF that are associated with specific acquisition devices. Using this information, we controlled FALM $L_f$ values for imaging device $d$ to generate controlled $L_{f,d}$ values by subtracting the average FALM values within each device $\mu_{f,d}$ and adding the grand average face image lightness $\mu_f$.

The ``Corrected'' dataset consists of only Enrollment Images (see Section~\ref{subsection:sources}).  In addition to being collected by a single acquisition device, these images are captured in front of a neutral grey background.  Consequently, they can be corrected for background image lightness.  This correction was performed by subtracting background lightness $L_{b,d}$ from FALM $L_{f,d}$ and reconstituting with the average difference between face and background lightness.  Table~\ref{table:datasets} (rows i - vi) shows the six datasets of FALM values from images used in this study.

\subsection{Ground-truth FALMs from calibrated equipment}
\label{subsection:colormeter}

\begin{figure}[ht]
    \centering
	\includegraphics[width=4.25cm]{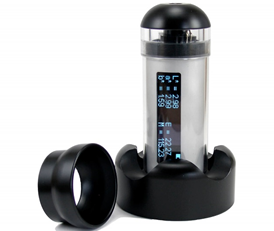}
    \caption{The DSM III Colormeter, Cortex Technology}
	\label{figure:colormeter}
\end{figure}

As part of~\cite{howard2019results}, ground-truth FALMs were recorded using a calibrated hand-held sensor (DSM III Colormeter, Cortex Technology, Figure~\ref{figure:colormeter}). The sensor measures skin color using an $RGB$ sensor to image a $7 mm^2$ patch of skin under standard illumination provided by two white light emitting diodes.  The device can accurately measure the color as well as erythema and melanin content of skin~\cite{clarys2000skin, diffey1984portable}.

For each of the 345 subjects in~\cite{howard2019results}, two bilateral measurements were collected by placing the colormeter on each subject's face approximately on the subject's zygomatic arch.  The two $sRGB$ measurements were collected in close succession and converted to the $L^{*}a^{*}b^{*}$ colorspace using the D65 illuminant $(L_{rc}$ and $L_{lc})$.  The subjects' skin was not cleaned prior to collection. As such, the colormeter measures are likely related to subjects' facultative pigmentation as well as any contributions from makeup in a manner similar to subjects' face images from cameras. The skin contacting surfaces of the colormeter were wiped with rubbing alcohol between subjects and the device itself was calibrated twice a day using a standardized procedure involving a white calibration plate provided by the colormeter manufacturer.  We verified that our ground-truth FALM readings matched skin tone readings reported in prior work~\cite{van2013objective,lee2019investigating}.  It's important to note that colormeter readings were collected on the same day as the images in the Enrollment and Acquisition image set and are from the same test subjects.  Images in the Historic image set are also from these same test subjects, but were collected on days prior to the day when colormeter readings were taken (see Section~\ref{subsection:sources}).

\subsection{Self-reported Fitzpatrick Skin Type}
\label{subsection:fst}

\begin{figure*}[ht]
    \centering
	\includegraphics[width=17cm]{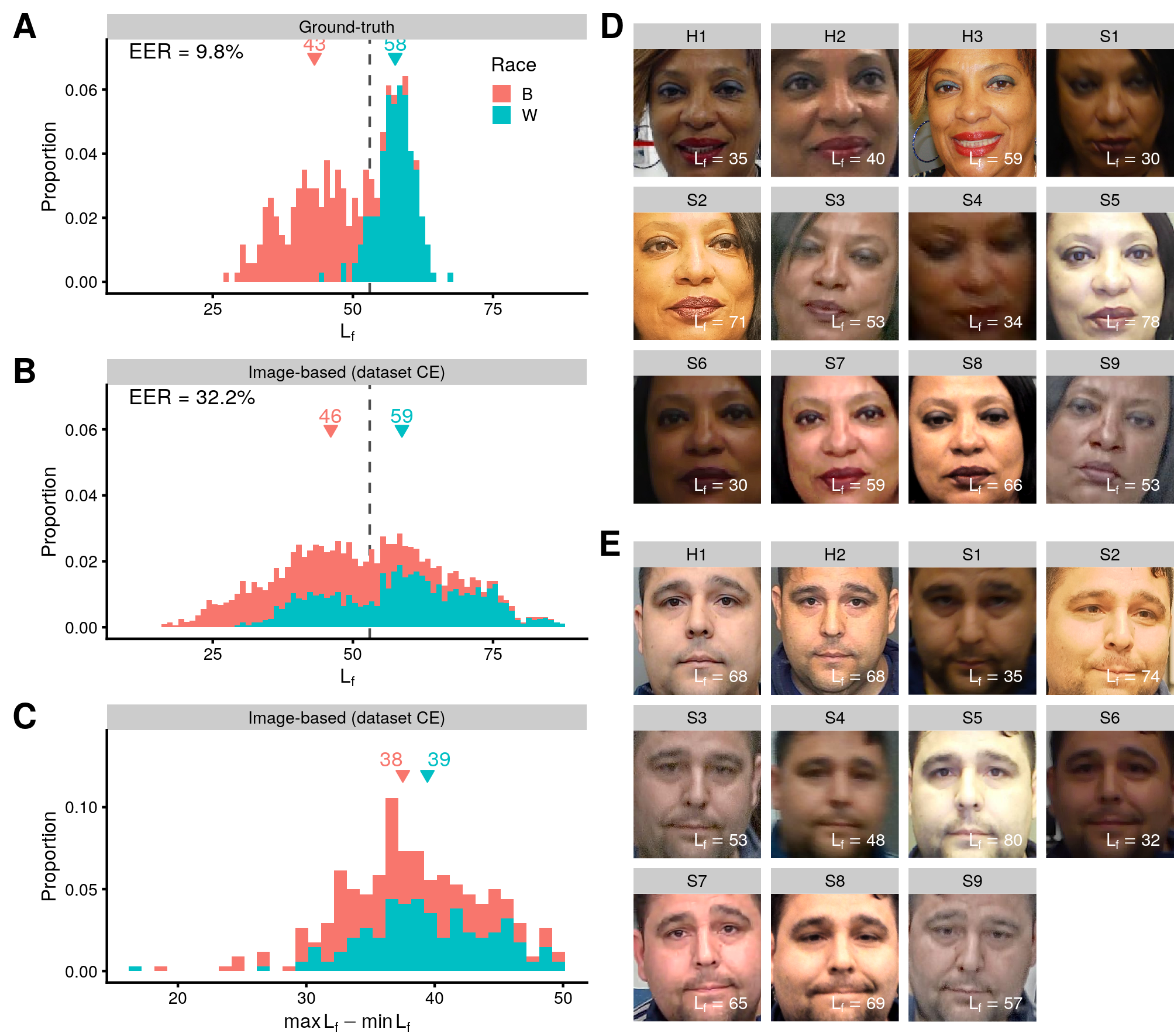}
    \caption{Distribution of face area lightness measures (FALMs) as measured by ground-truth colormeter device and as measured from images. \textbf{A.} Distribution of ground-truth FALMs ($L_{f}$) using the colormeter device.  Note bi-modal distribution with apparently distinct peaks for each self-reported race category. \textbf{B.} Distribution of $L_{f}$ values computed using images in the E dataset.  Note significant overlap between $L_{f}$ values for the two race categories.  \textbf{C.} Distribution of the range of $L_{f}$ values observed for each subject in the E dataset.  Note that $L_{f}$ value range frequently exceeds the $L_{f}$ difference between race categories. Dotted lines in A and B indicate equal error rate threshold. Triangles mark average within each race category. \textbf{D.} Sample images and computed $L_{f}$ values for a single subject self-identifying as Black or African-American. This subject's ground-truth colormeter $L_{f}$ was 28. \textbf{E.} Sample images and computed face lightness values for a single subject self-identifying as White.  This subject's ground-truth colormeter $L_{f}$ was 35. Note strong variation in $L_{f}$ values when measured from different images.  Facets labelled ``S'' are images taken by difference devices on the same day.  Facets labelled ``H'' are historic are images taken by different devices in different days.}
	\label{figure:Lightness}
\end{figure*}

Also as part of~\cite{howard2019results}, each subject self-reported their FST as part of a paper survey.  There are several ways to self-report FST, which vary in question, wording, the number of questions, and description of FST categories~\cite{eilers2013accuracy,he2014self,ash2015evaluation,sommers2019fitzpatrick}. The method we selected uses the single-question measure adapted from~\cite{eilers2013accuracy} because of its simplicity and because it includes specific descriptors that are more meaningful for darker skin tones.  Participants were asked ``Which of the following descriptions best matches your skin type?'' and allowed to select one option from a list that was most consistent with their experience.  Table~\ref{table:FST} shows the options provided and their mapping onto FST skin types, per~\cite{eilers2013accuracy}.  Responses were digitized for each test subject.

\begin{table}[ht]
    \centering
    \begin{tabular}{p{6cm}p{1cm}}
    \hline
    Option Text                                                             & FST\\
    \hline
    Highly sensitive, always burns, never tans                              & I  \\
    Very sun sensitive, burns easily, tans minimally                        & II \\
    Sun sensitive to skin, sometime burns, slowly tans to light brown       & III\\
    Minimally sun sensitive, burns minimally, always tans to moderate brown & IV \\
    Sun insensitive skin, rarely burns, tans well                           & V  \\
    Sun insensitive, never burns, deeply pigmented                          & VI \\
    \hline
    \end{tabular}
    \caption{FST question response options.}
	\label{table:FST}
\end{table}

\section{Results}
\label{section:Results}

\subsection{Variation in image-based FALMs}
\label{subsection:Variation}

Figure~\ref{figure:Lightness}A shows the distribution of ground-truth FALM values for the 181 Black and 164 White test subjects in the MdTF dataset (Table~\ref{table:datasets}, row vii).  We note little overlap between the two distributions and an equal error rate (EER) of 9.8\%, if a simple threshold based classification scheme were used.  Conversely, FALM values estimated from images where device and acquisition time varies (the CE dataset) were more broadly distributed (Figure~\ref{figure:Lightness}B) such that the distributions of FALMs between the two race categories overlapped to a greater extent (EER = 32.2\%)

This overlap was due to large variations in FALM values \textit{within subjects} when FALMs were taken from images.  Figure~\ref{figure:Lightness}C shows the range of FALM values for each individual in the CE dataset.  These intra-subject FALM values ranged, on average, by 38 units for Black or African American subjects and by 39 for White subjects, corresponding to more than a 2-fold difference in measured face area lightness, for a single subject, from image to image.  This variation across images is 3 times larger than the 13 point difference in the average FALM of individuals in the two race groups (Figure~\ref{figure:Lightness}B).

The large image-to-image variation in FALMs is demonstrated for two examples subjects in Figure~\ref{figure:Lightness}D-E.  The subject in Figure~\ref{figure:Lightness}D self-identified as Black or African-American with a ground-truth FALM of 28 and the subject in Figure~\ref{figure:Lightness}E self-identified as White with a ground-truth FALM reading of 35.  The variation in FALMs across images indicates that measuring face area lightness from uncontrolled images does not provide a suitable estimate of a subject's ground-truth FALM as determined by the colormeter instrument.

\subsection{Control in image-based FALMs}
\label{subsection:Control}

If images captured on various devices at various times are unsuitable for estimating ground-truth FALMs, what level of control must be added to image capture to allow for FALMs from images that approaches the ground-truth FALMs of the colormeter?  To answer this, we next examined how controlling certain factors during image acquisition impacts the image-to-image variation in image-based FALM.

\begin{figure}[ht]
    \centering
	\includegraphics[width=8.5cm]{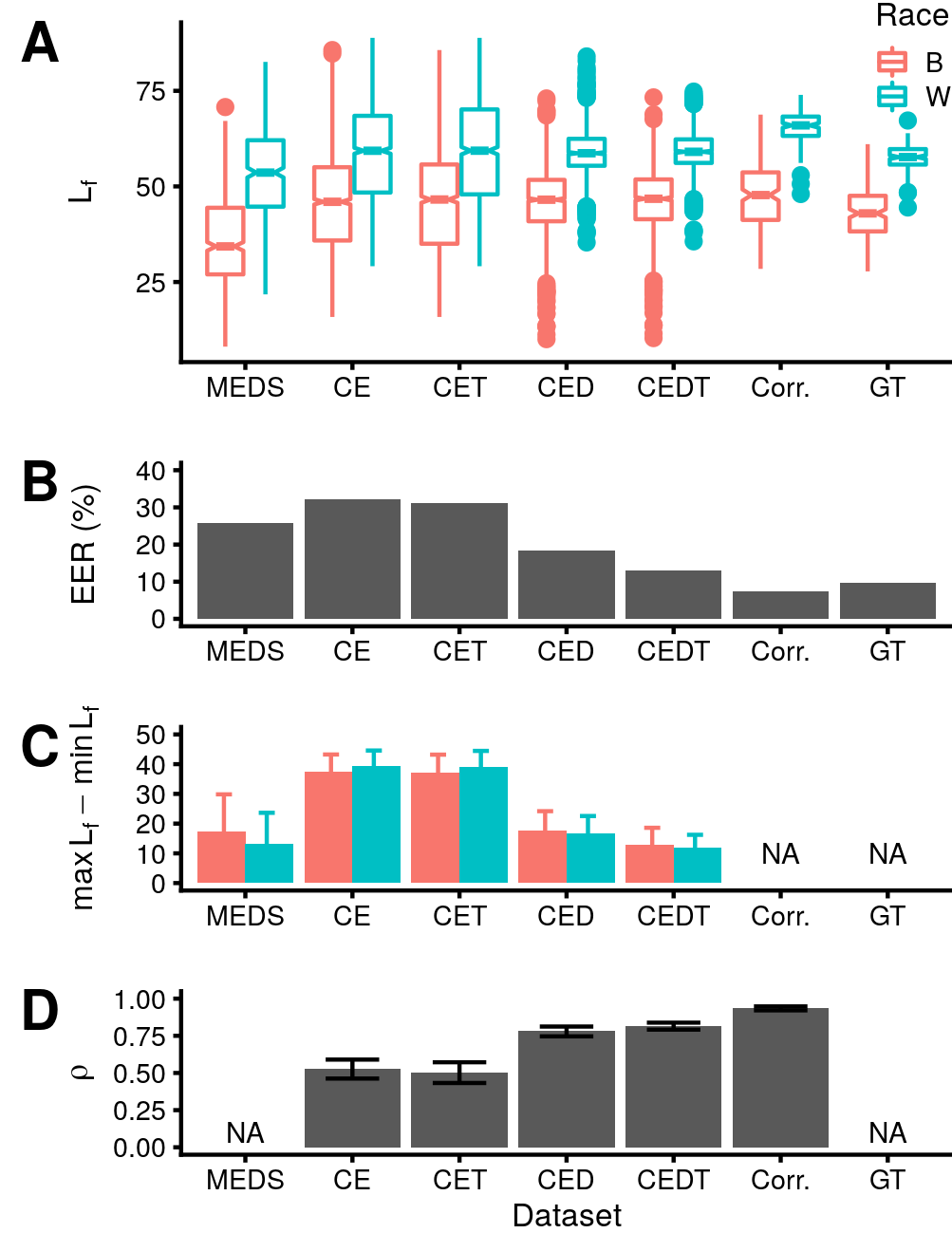}
    \caption{Variation in face area lightness measures (FALMs). \textbf{A.} Distributions of FALMs ($L_{f}$) across B and W subjects for each dataset. Equal error rate between FALM distributions for B and W subjects.  \textbf{C.} Range of intra subject $L_{f}$ for each dataset.  Error bars are standard deviation across subjects.  Note Corrected (Corr.) and Ground-truth (GT) datasets had only one image per subject so range could not be computed. \textbf{D.} Pearson correlation ($\rho$) between $L_{f}$ values from each dataset and the ground-truth colormeter $L_{f}$ values.  Error bars are 95\% confidence intervals.}
	\label{figure:RaceDist}
\end{figure}

Figure~\ref{figure:RaceDist}A shows the distributions FALMs from the datasets described in Table~\ref{table:datasets}.  As our measure of similarity to the ground-truth FALMs from the colormeter, we quantified the EER between FALM distributions across race categories in Figure~\ref{figure:RaceDist}B (recall from Figure~\ref{figure:Lightness} that the distributions of ground-truth FALMs disaggregated by race has an EER of 9.8\%).  As our measure of image-to-image variation within subject, we quantified the range of intra-subject FALM values for each dataset (Figure~\ref{figure:RaceDist}C).  The range of FALM values could not be computed for the Corrected (Corr.) or Ground-truth (GT) datasets because they had only one sample per subject.

When estimating FALMs from images, EER was highest when only environment was controlled (CE dataset, EER = 32\%) and lowest for the Corrected dataset (EER = 8\%).  The EER of 8\% for the Corrected dataset was comparable to the EER of ground-truth FALMs as measured by the colormeter (GT dataset, EER = 9.8\%).  The biggest single decline in EER and in the range of FALM values for each individual was observed when controlling for device (compare CED and CET in Figure~\ref{figure:RaceDist}B and C).  This suggests that variation across imaging devices is a major source of lightness variation when the images are acquired in a common environment.

In terms of EER, the MEDS dataset fell between the CET and CED datasets.  The average range of FALM values for MEDS images was lower than for CE or CET datasets.  However, MEDS images also had generally lower FALM values for subjects in both race categories (Figure~\ref{figure:RaceDist}A).  Overall, observations for the MEDS dataset are in line with those from the datasets based on images from the MdTF and suggests caution when using FALM from images, gathered without strict controls and corrections, as a phenotypic measure.

Finally, we measured the correlation between FALM values estimated from images and ground-truth FALMs quantified by the colormeter (Figure~\ref{figure:RaceDist}D).  Correlation could not be estimated for the MEDS dataset because it had distinct subjects.  The correlation for the CE dataset was poor (dataset CE, Pearson's $\rho = 0.45$).  However, correlation improved when controlling for acquisition device and time (dataset EDT, Pearson's $\rho = 0.78$).  Correlation was highest for the Corrected dataset comprised of FALM values from images taken on a single device, under controlled conditions, with correction for neutral grey background (dataset Corr., Pearson's $\rho = 0.92$).  This indicates that, under controlled conditions, image-based FALM values are good estimates of ground-truth FALMs from the colormeter instrument.

\subsection{Lightness, race, and Fitzpatrick Skin Type}
\label{subsection:FitzToLightness}

\begin{figure}[ht]
    \centering
	\includegraphics[width=8.5cm]{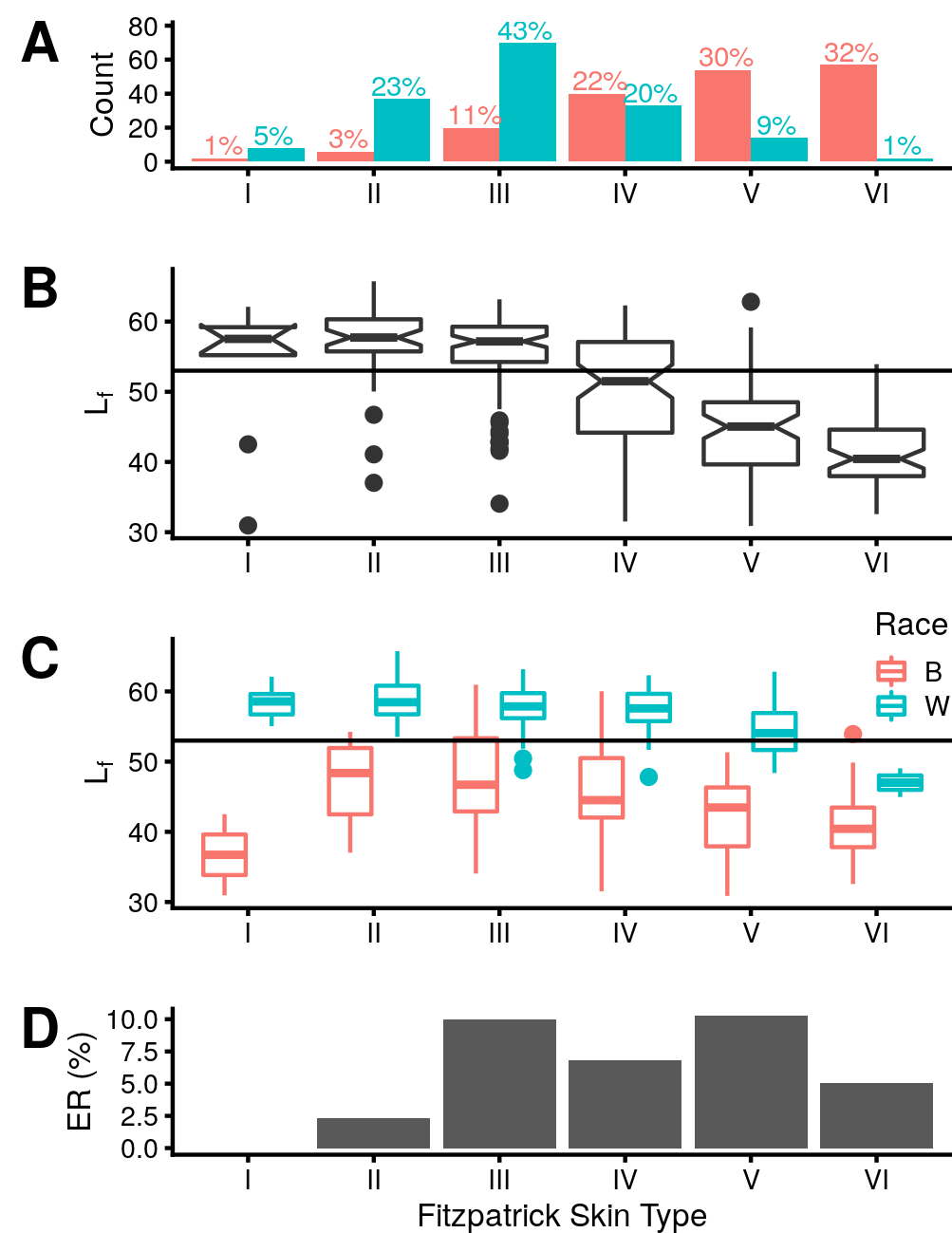}
    \caption{Relationship between Fitzpatrick Skin Type (FST) score and face area lightness measurements (FALMs). \textbf{A.} Distribution of self-reported FST by race. \textbf{B.} Distributions of FALM $(L_{f})$ values within each FST category.  Note apparent association between FST and $(L_{f})$. Horizontal line corresponds to overall equal error rate (EER) classification threshold. \textbf{C.} Distributions of $(L_{f})$ within each FST category by race. Note relatively smaller relationship between FST and $(L_{f})$ and relatively large separation between $(L_{f})$ distributions for each race within each FST category. \textbf{D.} Error rate (ER) values for race classification based on $(L_{f})$ within each FST category using the overall EER threshold.}
	\label{figure:FitzVsReflectance}
\end{figure}

We next examined the relationship between self-reported FST, self-reported race, and ground-truth FALMs from the colormeter.  Each subject assessed their own FST according to a standard scale (Table~\ref{table:FST}).  Figure~\ref{figure:FitzVsReflectance}A shows that FST was distributed differently when disaggregated by race ($\chi ^{2}(5) = 128.1, p << 0.001$).  Subjects that self-identified as Black or African-American chose FST VI most frequently whereas subjects that self-identified as White chose FST III as the most frequent category.

Intuitively, given that ground-truth FALM values also varied by race (Figure~\ref{figure:Lightness}A), we expected to observe a strong \textit{overall} association between ground-truth FALMs and FST.  This was confirmed in Figure~\ref{figure:FitzVsReflectance}B.   However, the apparent shift in FALM distributions observed in Figure~\ref{figure:FitzVsReflectance}B were actually due to different proportions of individuals from each group choosing each FST category (Figure~\ref{figure:FitzVsReflectance}A) while distributions of ground-truth FALMs within each race category remained largely invariant to FST (Figure~\ref{figure:FitzVsReflectance}C).  Indeed, using the EER threshold for the full population ($L_{f} = 52$, Figure~\ref{figure:RaceDist}A), there was little cross over between B and W ground-truth FALM distributions within each FST category.  This EER value peaked at only 10\% within FST III (Figure~\ref{figure:FitzVsReflectance}D), roughly equal to the whole group EER of 9.8\% from Figure~\ref{figure:Lightness}A.


Our conclusion is that FST is not a good predictor of ground-truth FALMs from the colormeter.  We measured the degree of association between FST, race, and ground-truth FALMs.  Correlation between FST and ground-truth FALMs (Kendall's $\tau = 0.51$) was lower than between race and ground-truth FALMs (Kendall's $\tau = 0.68$).  Bootstrap resampling showed the difference between these correlations to be significant ($\tau_{race}-\tau_{FST} = 0.17$, 95\% CI = 0.11-0.23).  Within each race category, the correlation between FST and ground-truth FALMs decreased (Kendall's $\tau = 0.23$) showing that most of the association between FST and ground-truth FALMs in our sample is due to the different proportions of subjects belonging to each race group choosing each FST category.

The relatively poor association between FST and ground-truth FALMs was confirmed by linear modelling of ground-truth FALMs with FST, which produced a poor fit ($L_{f} \sim FST$, $R^2 = 0.48$) relative to using race information alone ($L_{f} \sim race$), $R^2 = 0.72$).  Including both terms in the model hardly improved the fit over race alone ($L_{f} \sim FST + race$), $R^2 = 0.77$), although the full model fit was significantly better ($F(1)=429.68$, $p=2.2e^{-16}$).  This shows that race is actually a superior independent predictor of skin tone relative to self-reported FST, although FST does carry some additional information about skin tone.

\subsection{Impact of level of control on data interpretation}

\begin{figure*}[ht]
    \centering
	\includegraphics[width=15cm]{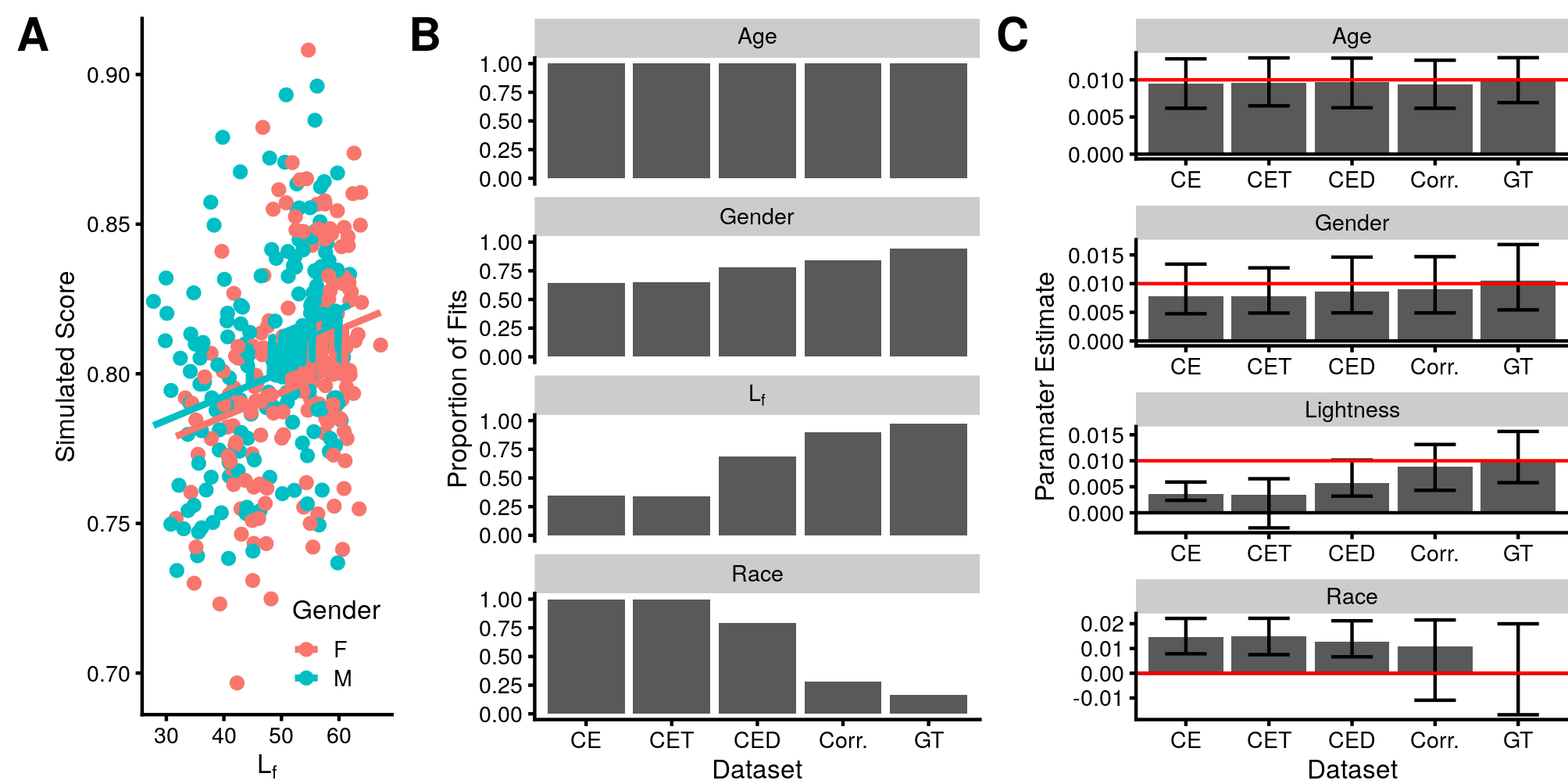}
    \caption{Model selection and parameter estimation is affected by level of control in phenotype estimation. \textbf{A.} Results of a simulated experiment with a known relationship whereby a simulated score is a function of face area lightness measure (FALM), gender, age (age not shown), and additive noise.  Critically, score is not a function of race. \textbf{B.} Proportion of times each demographic factor is selected based on model fits to resampled data FALM from different datasets.  Note that the likelihoods of erroneously selecting race (a type II error) and excluding FALM (a type I error) becomes greater with reduced level of control in teh dataset acquisition conditions. \textbf{C.} Parameter estimates for each demographic factor included in models fit to resampled data using different FALM estimates.  Red lines denote parameter estimates used to generate the simulated scores.  Note that the parameter estimate for FALM decreases with reduced level of control while the parameter estimate for race is increased. Error bars are 95\% bootstrap confidence intervals.}
	\label{figure:Modeling}
\end{figure*}

Poor phenotype estimation, such as measuring skin tone from uncontrolled images (Section~\ref{subsection:Variation} and~\ref{subsection:Control}) can have substantial impacts on experimental outcomes.  To illustrate this point, we executed a linear modelling experiment, which is a common way to analyze the relationship between demographic variables and biometric scores~\cite{beveridge2009factors, grother2013irex, cook2019demographic}.  We first generated simulated, mated, biometric similarity scores according to Equation~\ref{equation:modelsim}, for each subject in our dataset.  Note these simulated scores are based on subject $i$'s gender, age, and their ground-truth FALM as recorded by the colormeter.  To generate these scores, the intercept $\beta_0$ was set to 0.8, each continuous demographic variable was z-transformed, all effect sizes ($\beta_1$, $\beta_2$, and $\beta_3$) were set to 0.01, and the noise term was drawn from a normal distribution as $\epsilon \sim \mathcal{N}(\mu=0,\,\sigma=0.03)$.  Critically, these scores, visualized in Figure~\ref{figure:Modeling}A, are not a function of the subject's race.

\begin{equation}
  \label{equation:modelsim}
  S_{i,GT} \sim \beta_0 + \beta_1 gender_i + \beta_2 age_i + \beta_3 L_{f,a = GT,i} + \epsilon_i
\end{equation}

We then constructed a different model that allowed for the possibility of score being a function of a subject's race, as shown in Equation~\ref{equation:modelfit}.  This model used the FALM values from our different datasets $a$.  Estimation of model parameters $\beta$ was performed using ordinary least squares (OLS) to  fit 1,000 bootstrap replicates of the data.  Each replicate resampled 345 subjects from the population with replacement.  The simulated similarity score noise $\epsilon$ in Equation~\ref{equation:modelfit} was drawn separately for each replicate.  Also for each replicate, the optimal model was selected that minimizes the Akaike Information Criteria, $AIC = 2k - 2ln(\hat{L})$, where $k$ represents the number of estimated parameters in the model and $\hat{L}$ represents the maximum value of the model's fitted likelihood.  AIC measures the goodness of fit of the model while discouraging over-fitting with a penalty for increasing the number of model parameters $k$.  To find the optimal models, we used a step wise procedure in both directions.  This procedure resulted in a total 1,000 optimal models for each of our FALM datasets $a$.

\begin{equation}
\label{equation:modelfit}
    S_{i,a} \sim \beta_0 + \beta_1 gender_i + \beta_2 age_i + \beta_3 L_{f,a,i} + \\
    \beta_4 race_i + \epsilon_i
\end{equation}

When using ground-truth FALMs as measured by the colormeter as $L_{F}$ in Equation~\ref{equation:modelfit}, virtually all (97.2\%) of the optimal model fits included lightness and only 16.5\% made the type II error of including race (Figure~\ref{figure:Modeling}B).  The average parameter estimate for lightness in these models correctly reflected the strength of the simulated relationship.  However, in the models selecting race, the associated parameter estimate was, on average, negligible (Figure~\ref{figure:Modeling}C).  As expected, this shows that linear models that include ground-truth FALM values are very likely to indicate, correctly, that age, gender, and lightness are related to score, all with appropriate parameter estimates whereas the relationship between race and score is absent or negligible in most models.

On the other hand, models fit using FALM estimates from the poorly controlled CE dataset, led to a vastly different outcome.  Almost all (99.6\%) of these models included race in the optimal model as compared to only 29.2\% selecting lightless $L_{f}$ (Figure~\ref{figure:Modeling}B).  Further, the parameter estimate for $L_{f}$ in these models was far lower (0.003) than the true relationship between ground-truth FALMs and score we simulated (0.01, Figure~\ref{figure:Modeling}C).  This shows that linear models based on poorly measured FALM values are likely to lead to an incorrect interpretation of the relevant demographic factors, selecting age, gender, and race, but not measured lightness as related to score.  This over-estimates the impact of these demographic variables and entirely misses or under-estimates the impact of FALM on score.  Repeating this process for other datasets shows that the likelihood of a correct interpretation increases with increased level of acquisition control in the images from which FALMs are calculated (Figure~\ref{figure:Modeling}B-C). Thus, poor control in estimates of face phenotypes can lead to significant errors of interpretation regarding the significance of race categories in studies of biometric performance.

\section{Discussion}

In this study, we explored the feasibility of correctly quantifying an individual's face skin properties from a photograph.  We collected ground-truth face area lightness measures (FALMs) from a calibrated dermographic instrument, known as a colormeter.  We then compared these to FALMs assessed from an assortment of face photos.  We find that intra-subject FALMs assessed from photos can vary greatly from image to image, three times more than the average difference in ground-truth FALMs observed between the two race groups in our study (White and Black or African-American).  This intra-subject variation was present to similar degree in the NIST MEDS dataset commonly used in biometric performance assessment and is likely present in all computer vision datasets of humans where acquisition conditions are uncontrolled.  A measure that varies more within subject than it does between subject groups is a poor descriptor of the properties of the subject relative to the group.  We believe this is strong evidence that skin tone for use in evaluations of computer vision applications should not be ascertained from images captured in an uncontrolled environment or scraped off the web.

However, this study also shows that it is possible to obtain reliable estimates of skin tone from some images. Prior work has used face images acquired by a single device under constant conditions to measure relative skin reflectance after correcting for a neutral grey background present in the images~\cite{cook2019demographic}, and in this study we validate that approach.  FALMs estimated from images and using such corrections correlated strongly ($\rho = 0.92$) with ground-truth FALMs collected using the colormeter.  Thus, an accurate measurement of relative skin tone can be obtained even when a calibrated skin color meter is not available.

Next, the computer vision community has recently begun using Fitzpatrick Skin Type (FST) categories to describe skin tone in images, for the purpose of evaluating algorithms across this measure.  This methodology has been proposed in studies of gender classification~\cite{buolamwini2018gender}, biometric recognition~\cite{muthukumar2018understanding, krishnapriya2020issues, krishnapriya2021analysis, lu2019experimental}, and pedestrian detection~\cite{wilson2019predictive} algorithms.  It has also been suggested as a standardized method for documenting the performance of a generic machine learning algorithm~\cite{mitchell2019model}.  Our work shows that this novel use of FST may be problematic for at least three reasons.  First, as we discuss, FST was designed to classify UV sensitivity of an individual with specific labels assigned to each category.  FST is not an arbitrary ordinal scale and other ordinal scales with different category labels or a different method for arriving at these labels are not likely to produce equivalent results.  FST has been shown in medical literature to be a generally unreliable estimator of skin pigmentation~\cite{westerhof1990relation} and a specifically unreliable estimator for people of color~\cite{leenutaphong1995relationship, pichon2010measuring, galindo2007sun, sommers2019fitzpatrick}.  FST assessment is subject to inter-rater reliability issues~\cite{krishnapriya2021analysis} and known rater biases~\cite{harrison1999all, reeder2010questionnaire, hill2002race}, most notably conflating skin tone and other features related to the race of the subject and of the rater~\cite{ware2020racial}.  Because of these concerns, we believe FST is a poor choice for evaluating computer vision applications.

Second, in the medical literature, FST is arrived at by either self-report or physician accessed direct assessment.  Both require access to the physical subject for whom an FST measure is being calculated.  All existing computer vision work that has used FST measures has done so by having human raters judge the skin tone of subjects in images~\cite{buolamwini2018gender, lu2019experimental, krishnapriya2020issues, wilson2019predictive, krishnapriya2021analysis}.  However, as this study has shown, the face image lightness of the same subject varies greatly across uncontrolled images. Because of this assessment technique, we believe it is inaccurate to even describe the arrived at quantifications in~\cite{buolamwini2018gender, lu2019experimental, krishnapriya2020issues, wilson2019predictive, krishnapriya2021analysis} as Fitzpatrick Skin Types.  These studies have measured something using an image, but it was unlikely a good estimator of the FST phenotype, and is almost certainly not FST as the term is conceptualized in the medical community.

Third, even when FST types are calculated in manner supported by the medical literature, as we have done here, the six point self-reported FST is a poorer predictor of skin tone than even the binary race categories self-reported by the population in our study.  We have shown that the apparent aggregate relationship between FST and ground-truth FALMs is mainly due to different proportions of people in each race category selecting different FST values and a weak relationship between FST and ground-truth FALMs within each race category.  We believe this is strong evidence that a separate phenotypic measure should be used to assess skin properties in the assessment of computer vision algorithms generally and biometric performance in particular.

We summarize our findings with respect to Fitzpatrick Skin Types as follows: Don't use this measure to evaluate computer vision applications, it's unreliable, particularly with people of color. The medical community agrees with this assessment. In spite of this, if you choose to use FST classifications in an evaluation of computer vision applications, you may only arrive at FST determinations by in-person interview with a test subject. Other measures of ``FST'' from images of test subjects are prone to significant intra-subject, image-to-image variation in observed skin tone and are not, in fact, FST.  In general, when using an ordinal scale to classify skin properties, changing the survey instrument or changing the assigned categories changes the scale.  Care should be taken to explain how new scales were developed and validated before they are used in scientific studies. Finally, if you have chosen to use the FST ordinal scale and chosen to calculate it correctly, be aware this measure is a poor descriptor of skin tone and should not be used as such in evaluations of computer vision applications.

Lastly, we show that poor estimates of skin tone can lead to significant errors in interpretation of linear models relating demographic variables to biometric performance, a finding that is likely true of phenotypic measures in general.  In our study, ground-truth FALMs from the colormeter was strongly correlated with race.  When FALMs measured from images were used in a model fitting exercise, race replaced lightness in optimal models of simulated biometric performance even when simulated performance was not actually related to race.  This indicates that studies of demographic effects on performance should either \textit{a priori} determine which correlated variables (e.g. race or lightness) should be used in modelling or be cautious in their interpretation of the optimal model.  Minimizing error in measurement of phenotypes is necessary to avoid confusion between phenotypes and any correlated demographic groups.


\section*{Acknowledgments}
\label{section:Acknowledgements}

This research was sponsored by the Department of Homeland Security, Science and Technology Directorate on contract number W911NF-13-D-0006-0003. The views presented here are those of the authors and do not represent those of the Department of Homeland Security, the U.S. Government, or their employers.  The data were acquired using the IRB protocol ``Development and Evaluation of Enhanced Screening'' number 120180237, approved by New England IRB.

\ifCLASSOPTIONcaptionsoff
  \newpage
\fi



\bibliographystyle{IEEEtran}
\bibliography{tbiom-report}

\begin{thebibliography}{10}
\providecommand{\url}[1]{#1}
\csname url@samestyle\endcsname
\providecommand{\newblock}{\relax}
\providecommand{\bibinfo}[2]{#2}
\providecommand{\BIBentrySTDinterwordspacing}{\spaceskip=0pt\relax}
\providecommand{\BIBentryALTinterwordstretchfactor}{4}
\providecommand{\BIBentryALTinterwordspacing}{\spaceskip=\fontdimen2\font plus
\BIBentryALTinterwordstretchfactor\fontdimen3\font minus
  \fontdimen4\font\relax}
\providecommand{\BIBforeignlanguage}[2]{{%
\expandafter\ifx\csname l@#1\endcsname\relax
\typeout{** WARNING: IEEEtran.bst: No hyphenation pattern has been}%
\typeout{** loaded for the language `#1'. Using the pattern for}%
\typeout{** the default language instead.}%
\else
\language=\csname l@#1\endcsname
\fi
#2}}
\providecommand{\BIBdecl}{\relax}
\BIBdecl

\bibitem{suresh2019framework}
H.~Suresh and J.~V. Guttag, ``A framework for understanding unintended
  consequences of machine learning,'' \emph{arXiv preprint arXiv:1901.10002},
  2019.

\bibitem{howard2019effect}
J.~J. Howard, Y.~Sirotin, and A.~Vemury, ``The effect of broad and specific
  demographic homogeneity on the imposter distributions and false match rates
  in face recognition algorithm performance,'' in \emph{Proc. 10-th IEEE
  International Conference on Biometrics Theory, Applications and Systems,
  BTAS}, 2019.

\bibitem{vangara2019characterizing}
K.~Vangara, M.~C. King, V.~Albiero, K.~Bowyer \emph{et~al.}, ``Characterizing
  the variability in face recognition accuracy relative to race,'' in
  \emph{Proceedings of the IEEE Conference on Computer Vision and Pattern
  Recognition Workshops}, 2019, pp. 0--0.

\bibitem{grother2019face3}
P.~Grother, M.~Ngan, and K.~Hanaoka, ``{Face Recognition Vendor Test (FRVT)
  Part 3: Demographic Effects},'' Tech. Rep., 2019.

\bibitem{ma2018race}
D.~S. Ma, K.~Koltai, R.~M. McManus, A.~Bernhardt, J.~Correll, and
  B.~Wittenbrink, ``Race signaling features: Identifying markers of racial
  prototypicality among asians, blacks, latinos, and whites,'' \emph{Social
  Cognition}, vol.~36, no.~6, pp. 603--625, 2018.

\bibitem{cook2019demographic}
C.~M. Cook, J.~J. Howard, Y.~B. Sirotin, J.~L. Tipton, and A.~R. Vemury,
  ``Demographic effects in facial recognition and their dependence on image
  acquisition: An evaluation of eleven commercial systems,'' \emph{IEEE
  Transactions on Biometrics, Behavior, and Identity Science}, vol.~1, no.~1,
  pp. 32--41, 2019.

\bibitem{buolamwini2018gender}
J.~Buolamwini and T.~Gebru, ``Gender shades: Intersectional accuracy
  disparities in commercial gender classification,'' in \emph{Conference on
  fairness, accountability and transparency}, 2018, pp. 77--91.

\bibitem{muthukumar2018understanding}
V.~Muthukumar, T.~Pedapati, N.~Ratha, P.~Sattigeri, C.-W. Wu, B.~Kingsbury,
  A.~Kumar, S.~Thomas, A.~Mojsilovic, and K.~R. Varshney, ``Understanding
  unequal gender classification accuracy from face images,'' \emph{arXiv
  preprint arXiv:1812.00099}, 2018.

\bibitem{krishnapriya2020issues}
K.~Krishnapriya, V.~Albiero, K.~Vangara, M.~C. King, and K.~W. Bowyer, ``Issues
  related to face recognition accuracy varying based on race and skin tone,''
  \emph{IEEE Transactions on Technology and Society}, vol.~1, no.~1, pp. 8--20,
  2020.

\bibitem{lu2019experimental}
B.~Lu, J.-C. Chen, C.~D. Castillo, and R.~Chellappa, ``An experimental
  evaluation of covariates effects on unconstrained face verification,''
  \emph{IEEE Transactions on Biometrics, Behavior, and Identity Science},
  vol.~1, no.~1, pp. 42--55, 2019.

\bibitem{wilson2019predictive}
B.~Wilson, J.~Hoffman, and J.~Morgenstern, ``Predictive inequity in object
  detection,'' \emph{arXiv preprint arXiv:1902.11097}, 2019.

\bibitem{mitchell2019model}
M.~Mitchell, S.~Wu, A.~Zaldivar, P.~Barnes, L.~Vasserman, B.~Hutchinson,
  E.~Spitzer, I.~D. Raji, and T.~Gebru, ``Model cards for model reporting,'' in
  \emph{Proceedings of the conference on fairness, accountability, and
  transparency}, 2019, pp. 220--229.

\bibitem{harrison1999all}
S.~L. Harrison and P.~G. B{\"u}ttner, ``Do all fair-skinned caucasians consider
  themselves fair?'' \emph{Preventive medicine}, vol.~29, no.~5, pp. 349--354,
  1999.

\bibitem{reeder2010questionnaire}
A.~I. Reeder, V.~A. Hammond, and A.~R. Gray, ``Questionnaire items to assess
  skin color and erythemal sensitivity: reliability, validity, and ``the dark
  shift'','' \emph{Cancer Epidemiology and Prevention Biomarkers}, vol.~19,
  no.~5, pp. 1167--1173, 2010.

\bibitem{hill2002race}
M.~E. Hill, ``Race of the interviewer and perception of skin color: Evidence
  from the multi-city study of urban inequality,'' \emph{American Sociological
  Review}, pp. 99--108, 2002.

\bibitem{krishnapriya2021analysis}
K.~Krishnapriya, M.~C. King, and K.~W. Bowyer, ``Analysis of manual and
  automated skin tone assignments for face recognition applications,''
  \emph{arXiv e-prints}, pp. arXiv--2104, 2021.

\bibitem{westerhof1990relation}
W.~Westerhof, O.~Estevez-Uscanga, J.~Meens, A.~Kammeyer, M.~Durocq, and
  I.~Cario, ``The relation between constitutional skin color and
  photosensitivity estimated from uv-induced erythema and pigmentation
  dose-response curves.'' \emph{Journal of investigative dermatology}, vol.~94,
  no.~6, 1990.

\bibitem{leenutaphong1995relationship}
V.~Leenutaphong, ``Relationship between skin color and cutaneous response to
  ultraviolet radiation in thai,'' \emph{Photodermatology, photoimmunology \&
  photomedicine}, vol.~11, no. 5-6, pp. 198--203, 1995.

\bibitem{pichon2010measuring}
L.~C. Pichon, H.~Landrine, I.~Corral, Y.~Hao, J.~A. Mayer, and K.~D. Hoerster,
  ``Measuring skin cancer risk in african americans: is the fitzpatrick skin
  type classification scale culturally sensitive,'' \emph{Ethn Dis}, vol.~20,
  no.~2, pp. 174--179, 2010.

\bibitem{galindo2007sun}
G.~R. Galindo, J.~A. Mayer, D.~Slymen, D.~D. Almaguer, E.~Clapp, L.~C. Pichon,
  K.~Hoerster, and J.~P. Elder, ``Sun sensitivity in 5 us ethnoracial groups,''
  \emph{CUTIS-NEW YORK-}, vol.~80, no.~1, p.~25, 2007.

\bibitem{sommers2019fitzpatrick}
M.~S. Sommers, J.~D. Fargo, Y.~Regueira, K.~M. Brown, B.~L. Beacham, A.~R.
  Perfetti, J.~S. Everett, and D.~J. Margolis, ``Are the fitzpatrick skin
  phototypes valid for cancer risk assessment in a racially and ethnically
  diverse sample of women?'' \emph{Ethnicity \& disease}, vol.~29, no.~3, pp.
  505--512, 2019.

\bibitem{klare2012face}
B.~F. Klare, M.~J. Burge, J.~C. Klontz, R.~W.~V. Bruegge, and A.~K. Jain,
  ``Face recognition performance: Role of demographic information,'' \emph{IEEE
  Transactions on Information Forensics and Security}, vol.~7, no.~6, pp.
  1789--1801, 2012.

\bibitem{diffey1980ultraviolet}
B.~Diffey, ``Ultraviolet radiation physics and the skin,'' \emph{Physics in
  Medicine \& Biology}, vol.~25, no.~3, p. 405, 1980.

\bibitem{van2013objective}
M.~van~der Wal, M.~Bloemen, P.~Verhaegen, W.~Tuinebreijer, H.~de~Vet, P.~van
  Zuijlen, and E.~Middelkoop, ``Objective color measurements: clinimetric
  performance of three devices on normal skin and scar tissue,'' \emph{Journal
  of Burn Care \& Research}, vol.~34, no.~3, pp. e187--e194, 2013.

\bibitem{clarys2000skin}
P.~Clarys, K.~Alewaeters, R.~Lambrecht, and A.~Barel, ``Skin color
  measurements: comparison between three instruments: the
  chromameter{\textregistered}, the dermaspectrometer{\textregistered} and the
  mexameter{\textregistered},'' \emph{Skin research and technology}, vol.~6,
  no.~4, pp. 230--238, 2000.

\bibitem{weatherall1992skin}
I.~L. Weatherall and B.~D. Coombs, ``Skin color measurements in terms of cielab
  color space values,'' \emph{Journal of investigative dermatology}, vol.~99,
  no.~4, pp. 468--473, 1992.

\bibitem{nishidate2008visualizing}
I.~Nishidate, K.~Sasaoka, T.~Yuasa, K.~Niizeki, T.~Maeda, and Y.~Aizu,
  ``Visualizing of skin chromophore concentrations by use of rgb images,''
  \emph{Optics letters}, vol.~33, no.~19, pp. 2263--2265, 2008.

\bibitem{everett2012making}
J.~S. Everett, M.~Budescu, and M.~S. Sommers, ``Making sense of skin color in
  clinical care,'' \emph{Clinical nursing research}, vol.~21, no.~4, pp.
  495--516, 2012.

\bibitem{ware2020racial}
O.~R. Ware, J.~E. Dawson, M.~M. Shinohara, and S.~C. Taylor, ``Racial
  limitations of fitzpatrick skin type.'' \emph{Cutis}, vol. 105, no.~2, pp.
  77--80, 2020.

\bibitem{fitzpatrick1988validity}
T.~B. Fitzpatrick, ``The validity and practicality of sun-reactive skin types i
  through vi,'' \emph{Archives of dermatology}, vol. 124, no.~6, pp. 869--871,
  1988.

\bibitem{he2014self}
S.~Y. He, C.~E. McCulloch, W.~J. Boscardin, M.-M. Chren, E.~Linos, and S.~T.
  Arron, ``Self-reported pigmentary phenotypes and race are significant but
  incomplete predictors of fitzpatrick skin phototype in an ethnically diverse
  population,'' \emph{Journal of the American Academy of Dermatology}, vol.~71,
  no.~4, pp. 731--737, 2014.

\bibitem{chan2005assessing}
J.~L. Chan, A.~Ehrlich, R.~C. Lawrence, A.~N. Moshell, M.~L. Turner, and A.~B.
  Kimball, ``Assessing the role of race in quantitative measures of skin
  pigmentation and clinical assessments of photosensitivity,'' \emph{Journal of
  the American Academy of Dermatology}, vol.~52, no.~4, pp. 609--615, 2005.

\bibitem{ash2015evaluation}
C.~Ash, G.~Town, P.~Bjerring, and S.~Webster, ``Evaluation of a novel skin tone
  meter and the correlation between fitzpatrick skin type and skin color,''
  \emph{Photonics \& Lasers in Medicine}, vol.~4, no.~2, pp. 177--186, 2015.

\bibitem{levin2006distortions}
D.~T. Levin and M.~R. Banaji, ``Distortions in the perceived lightness of
  faces: the role of race categories.'' \emph{Journal of Experimental
  Psychology: General}, vol. 135, no.~4, p. 501, 2006.

\bibitem{hayward2013other}
W.~G. Hayward, K.~Crookes, and G.~Rhodes, ``The other-race effect: Holistic
  coding differences and beyond,'' \emph{Visual Cognition}, vol.~21, no. 9-10,
  pp. 1224--1247, 2013.

\bibitem{eilers2013accuracy}
S.~Eilers, D.~Q. Bach, R.~Gaber, H.~Blatt, Y.~Guevara, K.~Nitsche, R.~V. Kundu,
  and J.~K. Robinson, ``Accuracy of self-report in assessing fitzpatrick skin
  phototypes i through vi,'' \emph{JAMA dermatology}, vol. 149, no.~11, pp.
  1289--1294, 2013.

\bibitem{whitelam2017iarpa}
C.~Whitelam, E.~Taborsky, A.~Blanton, B.~Maze, J.~Adams, T.~Miller, N.~Kalka,
  A.~K. Jain, J.~A. Duncan, K.~Allen \emph{et~al.}, ``Iarpa janus benchmark-b
  face dataset,'' in \emph{proceedings of the IEEE conference on computer
  vision and pattern recognition workshops}, 2017, pp. 90--98.

\bibitem{howard2019results}
J.~Hasselgren, J.~J. Howard, Y.~Sirotin, J.~Tipton, and A.~Vemury, ``A scenario
  evaluation of high-throughput face biometric systems: Select results from the
  2019 department of homeland security biometric technology rally,'' \emph{MdTF
  Preprint}, 2020.

\bibitem{founds2011nist}
A.~P. Founds, N.~Orlans, W.~Genevieve, and C.~I. Watson, ``Nist special databse
  32-multiple encounter dataset ii (meds-ii),'' Tech. Rep., 2011.

\bibitem{taylor2014adaptive}
M.~J. Taylor and T.~Morris, ``Adaptive skin segmentation via feature-based face
  detection,'' in \emph{Real-Time Image and Video Processing 2014}, vol.
  9139.\hskip 1em plus 0.5em minus 0.4em\relax International Society for Optics
  and Photonics, 2014, p. 91390P.

\bibitem{diffey1984portable}
B.~Diffey, R.~Oliver, and P.~Farr, ``A portable instrument for quantifying
  erythema induced by ultraviolet radiation,'' \emph{British Journal of
  Dermatology}, vol. 111, no.~6, pp. 663--672, 1984.

\bibitem{lee2019investigating}
K.~C. Lee, A.~Bamford, F.~Gardiner, A.~Agovino, B.~Ter~Horst, J.~Bishop,
  A.~Sitch, L.~Grover, A.~Logan, and N.~Moiemen, ``Investigating the intra-and
  inter-rater reliability of a panel of subjective and objective burn scar
  measurement tools,'' \emph{Burns}, vol.~45, no.~6, pp. 1311--1324, 2019.

\bibitem{beveridge2009factors}
J.~R. Beveridge, G.~H. Givens, P.~J. Phillips, and B.~A. Draper, ``Factors that
  influence algorithm performance in the face recognition grand challenge,''
  \emph{Computer Vision and Image Understanding}, vol. 113, no.~6, pp.
  750--762, 2009.

\bibitem{grother2013irex}
P.~J. Grother, J.~R. Matey, E.~Tabassi, G.~W. Quinn, and M.~Chumakov, ``Irex
  vi-temporal stability of iris recognition accuracy,'' Tech. Rep., 2013.

\end{thebibliography}


\begin{IEEEbiography}[{\includegraphics[width=1in,height=1.25in,clip,keepaspectratio]{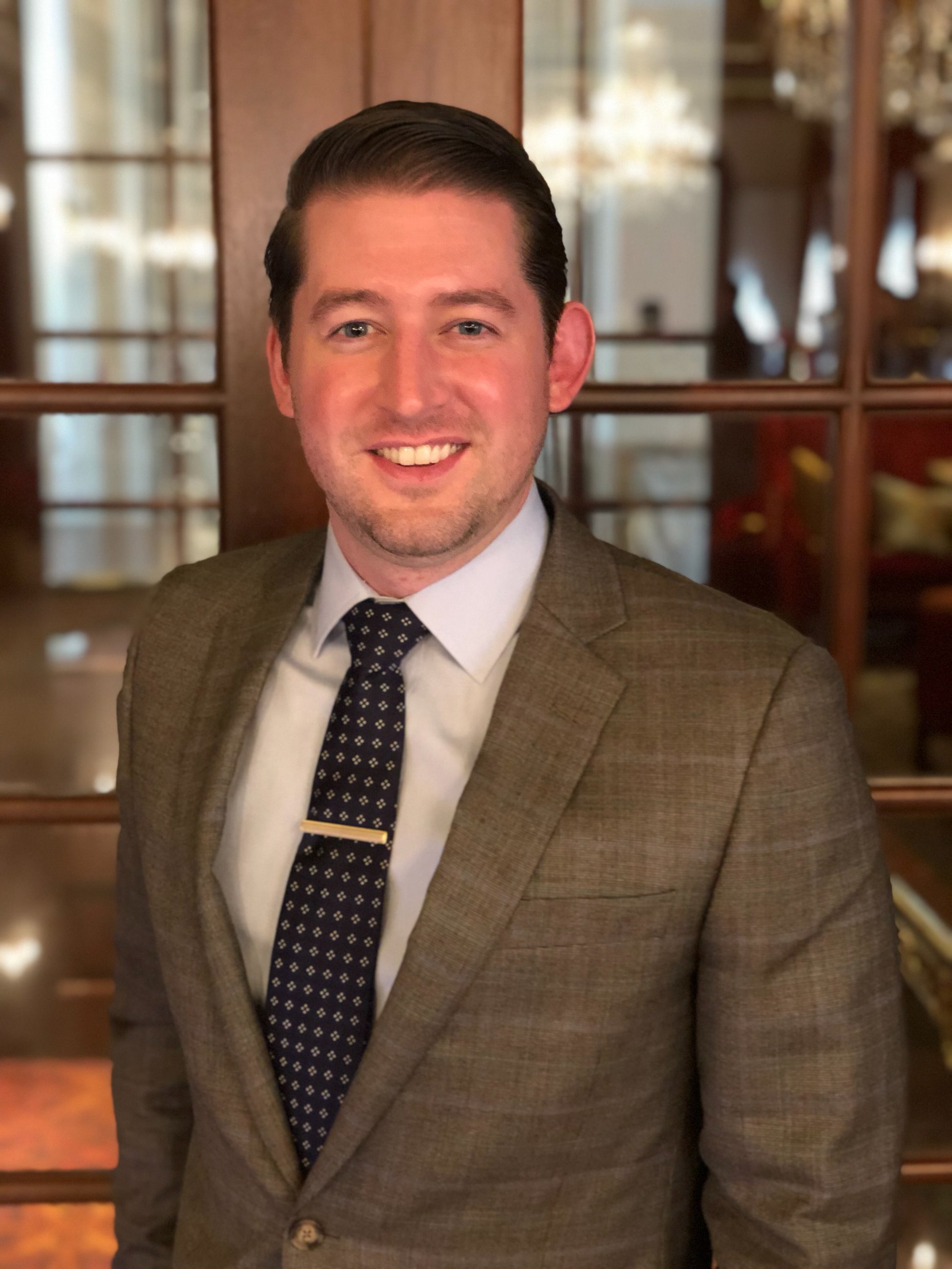}}]{John Howard}
Dr. Howard earned his Ph.D. in Computer Science from Southern Methodist University.  His thesis was on pattern recognition models for identifying subject specific match probability.  His current research interests include biometrics, computer vision, machine learning, testing human machine interfaces, pattern recognition, and statistics.  He has served as the principal investigator on numerous R\&D efforts across the intelligence community, Department of Defense, and other United States Government agencies.  He is currently the Principal Data Scientist at the Identity and Data Sciences Laboratory at SAIC where he conducts biometrics related research at the Maryland Test Facility.
\end{IEEEbiography}

\begin{IEEEbiography}[{\includegraphics[width=1in,height=1.25in,clip,keepaspectratio]{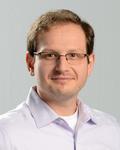}}]{Yevgeniy Sirotin}
Dr. Sirotin holds a Ph.D. in Neurobiology and Behavior from Columbia University and has diverse research interests in behavior and human computer interaction.  His past research spans mathematical psychology (cognitive modeling), neurophysiology (multi-spectral imaging of the brain), psychometrics (mechanisms of visual and olfactory perception), biometrics (design and testing of identity systems), and human factors (usability).  He currently works as Principal Investigator and Manager of the Identity and Data Sciences Laboratory at SAIC which supports applied research in biometric identity technologies at the Maryland Test Facility.
\end{IEEEbiography}

\begin{IEEEbiography}[{\includegraphics[width=1in,height=1.25in,clip,keepaspectratio]{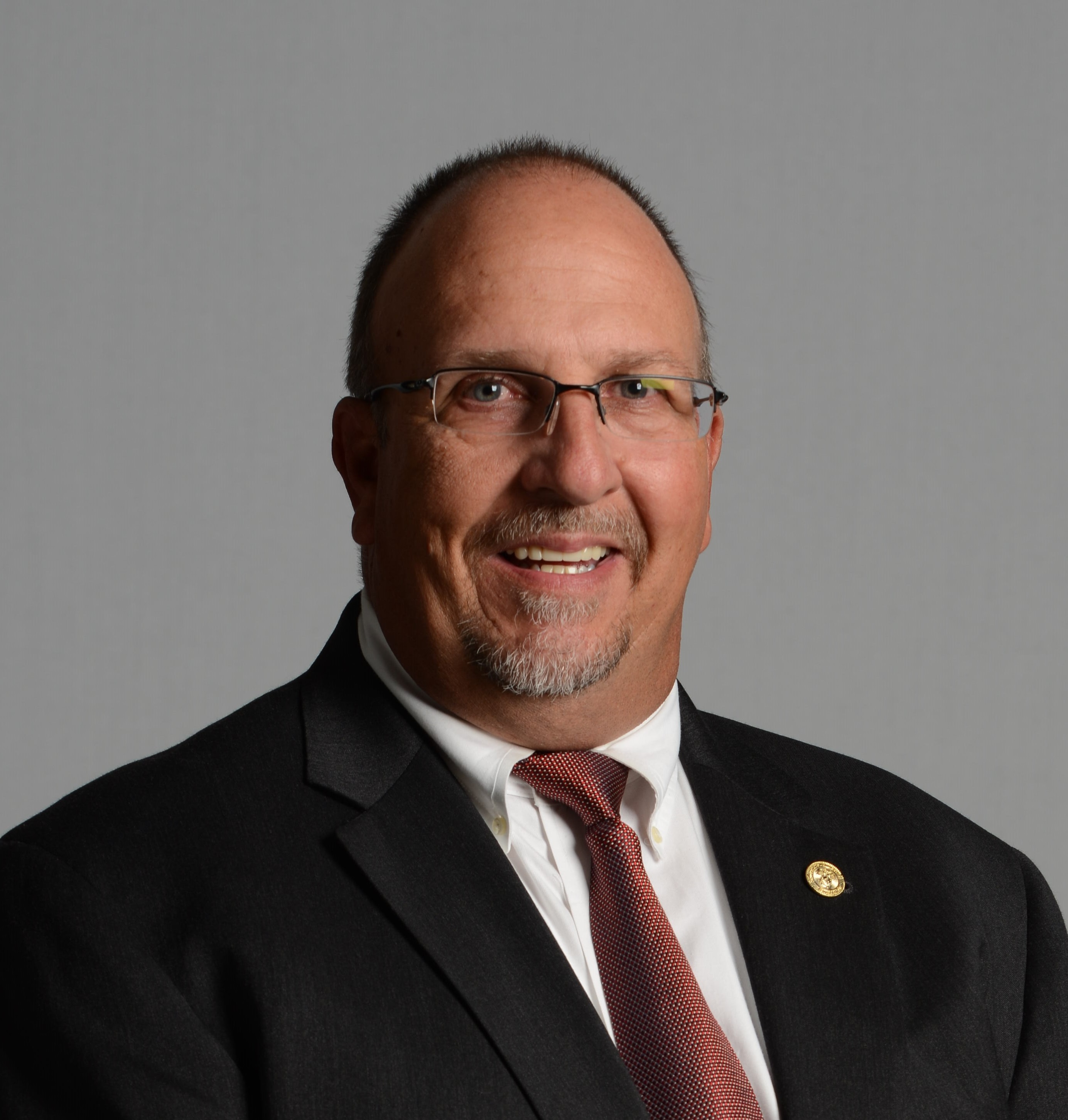}}]{Jerry Tipton}
Jerry Tipton is the Program Manager at SAIC's Identity and Data Sciences Lab.  He has over 20 years experience in the biometric industry with over 15 years managing research portfolios in support of various United States Government agencies.  He currently supports the Department of Homeland Security, Science and Technology Directorate at the Maryland Test Facility.
\end{IEEEbiography}

\begin{IEEEbiography}[{\includegraphics[width=1in,height=1.25in,clip,keepaspectratio]{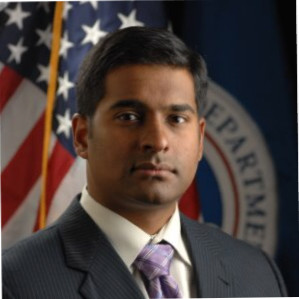}}]{Arun Vemury}
Arun Vemury received his M.S. in Computer Engineering from George Washington University.  His current research interests include biometrics, pattern recognition, machine learning, and operations research. He serves as the Director of the Biometrics and Identity Technology Center for the United States Department of Homeland Security Science and Technology Directorate.
\end{IEEEbiography}


\vfill




\end{document}